%% file: main.tex
\documentclass{article}

\usepackage{iclr2027_conference,times}

\usepackage[utf8]{inputenc} 
\usepackage[T1]{fontenc}    
\usepackage{url}            
\usepackage{booktabs}       
\usepackage{amsfonts}       
\usepackage{nicefrac}       
\usepackage{microtype}      
\usepackage{xcolor}         

\usepackage{graphicx}
\usepackage{subcaption}
\usepackage{multirow}
\usepackage{algorithm}
\usepackage{algorithmic}
\usepackage{enumitem}
\usepackage{amsmath}
\usepackage{hyperref}       
\usepackage[capitalize,noabbrev,nameinlink]{cleveref}
\usepackage{tabularx}
\usepackage{wrapfig}
\definecolor{CiteBlue}{rgb}{0.000, 0.350, 0.700}  
\definecolor{teal}{rgb}{0.040, 0.470, 0.390}
\hypersetup{colorlinks, citecolor=teal, linkcolor=teal}
\usepackage[table]{xcolor}

\title{Concept Unlearning via Cross-Attention\\Activation Subspaces in Diffusion Models}

\iclrfinalcopy

\author{%
  Saemi Moon\textsuperscript{1} \quad
  Suhyeon Jun\textsuperscript{1} \quad
  Seoyeon Lee\textsuperscript{2} \quad
  Dongwoo Kim\textsuperscript{1,2}
  \\[0.5em]
  \textsuperscript{1}CSE, POSTECH  \quad
  \textsuperscript{2}GSAI, POSTECH
  \\[0.5em]
  \texttt{\{saemi, suhyeonjun, seoyeon26, dongwoo.kim\}@postech.ac.kr}
}

\usepackage{twemojis}
\usepackage{duckuments}

\newcommand{\figpla}[1][.5\textwidth]{\includegraphics[width=#1]{example-image-duck}}
\input{math_commands}
\input{premble}

\begin{document}

\maketitle
\lhead{Preprint}

\begin{abstract}
\input{sections/0_abstract}
\end{abstract}

\input{sections/1_introduction}
\input{sections/2_related_work}
\input{sections/3_method}

\input{sections/4_experiments}

\input{sections/5_conclusion}

\bibliography{references}
\bibliographystyle{iclr2027_conference}

\clearpage

\appendix
\crefalias{section}{appendix}
\crefalias{subsection}{appendix}

\input{sections/A_appendix}


\end{document}

%% file: math_commands.tex
\usepackage{amsmath,amsfonts,bm}

\def\eqref#1{equation~\ref{#1}}

\def\1{\bm{1}}

\def\mI{{\bm{I}}}

\DeclareMathAlphabet{\mathsfit}{\encodingdefault}{\sfdefault}{m}{sl}
\SetMathAlphabet{\mathsfit}{bold}{\encodingdefault}{\sfdefault}{bx}{n}

\newcommand{\R}{\mathbb{R}}

\newcommand{\softmax}{\mathrm{softmax}}

\newcommand{\Wk}{W_K}
\newcommand{\Wv}{W_V}

\newcommand{\Af}{\mathcal{A}_f}
\newcommand{\Ar}{\mathcal{A}_r}
\newcommand{\Pf}{P_F}

\newcommand{\clip}{\mathrm{CLIP}}

%% file: premble.tex
\newcommand{\red}[1]{{\color{red}#1}}

\newcommand{\sm}[1]{\textbf{\color{magenta}[SM: #1]}}

\newcommand{\ESD}{$\mathtt{ESD}$}
\newcommand{\Receler}{$\mathtt{Receler}$}
\newcommand{\Mace}{$\mathtt{MACE}$}
\newcommand{\SD}{$\mathtt{SD}$}
\newcommand{\UCE}{$\mathtt{UCE}$}
\newcommand{\CURE}{$\mathtt{CURE}$}
\newcommand{\AdaVD}{$\mathtt{AdaVD}$}
\newcommand{\SAFREE}{$\mathtt{SAFREE}$}
\newcommand{\Ours}{$\mathtt{CASE}$}

\newcommand{\Celebrity}{$\mathtt{Celebrity}$}
\newcommand{\Style}{$\mathtt{Style}$}
\newcommand{\IP}{$\mathtt{IP}$}
\newcommand{\NSFW}{$\mathtt{NSFW}$}

\newcommand{\prompt}[1]{``\textit{#1}''}
\newcommand{\std}[1]{\pm\text{\scriptsize #1}}

%% file: sections/0_abstract.tex
Existing closed-form methods for concept unlearning in text-to-image diffusion models typically derive editing directions from fixed text embeddings, which may not fully capture how concepts are expressed across latent states, timesteps, and layers.
To capture this variation, we investigate cross-attention activations collected during denoising. In controlled probing experiments using the same anchor prompts, activation-derived bases achieve approximately five times the recall of text-derived bases on held-out prompts expressing the target concepts. Based on this finding, we propose Cross-Attention Subspace Erasure (\Ours{}), a closed-form method that constructs layer-specific forget and retain subspaces from cross-attention activations. These subspaces define a retain-constrained linear operator incorporated directly into cross-attention weights, requiring no gradient-based fine-tuning or additional inference-time computation. Across ten concepts spanning four categories, \Ours{} achieves the highest harmonic-mean score among evaluated baselines in all four categories, balancing suppression, retention, adversarial robustness, and generation quality. Further experiments demonstrate robustness to recovery attacks and a favorable suppression-retention trade-off when jointly unlearning up to 100 artistic styles. The benefits of activation-derived editing also extend to larger diffusion models, including SDXL and FLUX.

%% file: sections/1_introduction.tex
\section{Introduction}
\label{sec:intro}

Text-to-image diffusion models~\citep{rombach2022sd} have become standard tools for synthesizing photorealistic images from natural-language prompts~\citep{podell2024sdxl,saharia2022photorealistic,ramesh2022hierarchical}. Trained on web-scale data, these models can also reproduce content that a deployed system may need to restrict, including specific artistic styles, depictions of public figures, and unsafe imagery~\citep{schuhmann2022laion5b,qu2023unsafe,kumari2023conceptablation}. Concept unlearning aims to remove such concepts from a pretrained model without retraining it from scratch~\citep{gandikota2023esd,lu2024mace,biswas2025cure,wang2025ace}.

Concept unlearning requires both strong target suppression and selective preservation. The target concept should remain suppressed across diverse prompt formulations, including paraphrases and adversarial inputs. At the same time, the model's behavior on non-target concepts should remain close to that of the original model. These two objectives are inherently coupled because an edit that aggressively suppresses the target can also interfere with semantically related concepts. A practical unlearning method must therefore achieve a favorable balance between forgetting and retention.


Recent closed-form methods~\citep{gandikota2023uce,biswas2025cure,orgad2023time, gong2024rece, wang2025ace} address this trade-off through a single deterministic weight update without gradient-based fine-tuning. However, a common design choice among closed-form methods is to derive editing directions from prompt-level text representations, which remain fixed throughout denoising. Cross-attention activations, in contrast, depend on the evolving latent state, timestep, and layer. As a result, a text-derived editing basis may fail to cover the different ways in which the same concept is expressed during generation.

This observation motivates a simple question: should the editing basis be constructed from the cross-attention activations observed during denoising? Using the same anchor prompts, we compare bases formed from text-encoder representations and from cross-attention activations collected during denoising. A controlled probing experiment shows that the activation basis achieves approximately five times the average recall of the text basis across the \Style{}, \IP{}, and \Celebrity{} categories (\Cref{fig:binary-probe-natural}). This suggests that activation-space representations generalize better to unseen prompt formulations under the same anchor supervision. Unlike text embeddings, cross-attention activations incorporate the current image-side state and therefore characterize how the conditioning signal is expressed within the denoising computation.

\begin{figure}[t]
  \centering
  \includegraphics[width=\linewidth]{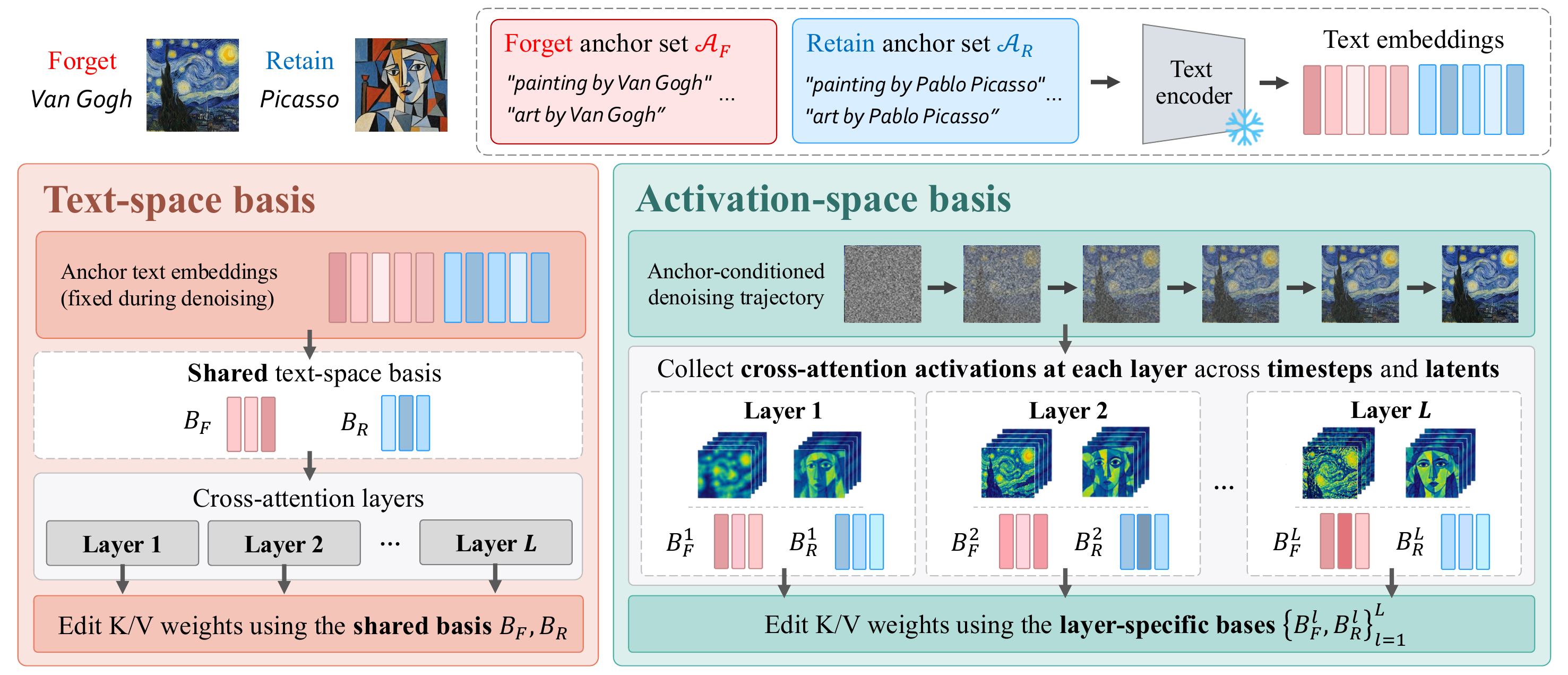}
  \caption{Comparison between text and activation bases. Prior closed-form methods build a shared text-space basis from anchor prompt embeddings. \Ours{} instead captures layer-specific cross-attention activations during denoising and constructs an activation-space basis for each cross-attention layer.}
  \label{fig:compare-basis}
\end{figure}

Motivated by this finding, we propose \Ours{}~(Cross-Attention Subspace Erasure), a closed-form unlearning method that constructs the editing basis directly from cross-attention activations (\Cref{fig:compare-basis}). \Ours{} traces short denoising trajectories for a small set of anchor prompts, collects activations at each cross-attention layer, and applies SVD to obtain layer-specific forget and retain subspaces. These subspaces define a closed-form linear operator that is folded directly into the key and value projection weights. The resulting edit requires no gradient-based fine-tuning and introduces no additional computation at inference time.

We evaluate \Ours{} on ten concepts spanning \Style{}, \IP{}, \Celebrity{}, and \NSFW{} categories in the Holistic Unlearning Benchmark (HUB)~\citep{moon2025holistic}. Across training-based, inference-time, and closed-form baselines, \Ours{} achieves the highest harmonic-mean score in all four categories, providing the strongest overall balance across target suppression, retention, adversarial robustness, and generation quality. We further find that the activation-space edit remains effective as the forget and retain anchor sets vary, performs competitively across diverse adversarial attacks, scales to multi-concept unlearning, and extends to SDXL and FLUX.1-dev.

%% file: sections/2_related_work.tex
\section{Related Work}
\label{sec:related}

\vspace{1mm}\paragraph{Closed-form concept unlearning.}
A growing line of work edits a text-to-image diffusion model in a single deterministic step, without gradient-based fine-tuning~\citep{orgad2023time, gong2024rece, wang2025ace}. \UCE~\citep{gandikota2023uce} solves for a closed-form update to the cross-attention key and value projections that redirects forget embeddings to a general concept while protecting a retain set. \CURE~\citep{biswas2025cure} replaces anchor-redirection with a spectral construction: it builds a forget basis from text-encoder embeddings of anchor prompts, applies a saturating spectral reweighting, and projects the resulting subspace out of the same weights. Both build the forget basis in text-embedding space. Like \CURE, our method combines forget and retain subspaces in a closed-form edit. We construct these subspaces from layer-specific cross-attention activations collected during denoising.

\paragraph{Inference-time concept unlearning.}
A separate family of methods intervenes only at inference rather than modifying model weights~\citep{schramowski2023sld,jain2024trasce,li2023seot,gaintseva2026casteer}. \SAFREE~\citep{yoon2025safree} projects prompt embeddings out of a toxic-concept subspace in text-embedding space and re-weights attention in the latent space. \AdaVD~\citep{Wang2025AdaVD} projects the value activations of each cross-attention layer onto the orthogonal complement of the target concept, with an adaptive shift that balances erasure against prior preservation. These methods adapt instantly to new concepts, but the intervention must be re-applied for every input and they represent the target through prompt-determined features. Our method instead folds a per-layer activation-space projection into the key and value weights once, adding no inference-time overhead.

\paragraph{Training-based concept unlearning.}
A complementary line erases concepts by gradient optimization~\citep{kumari2023conceptablation, heng2023sa,  lyu2024spm, fan2024salun, zhang2024advunlearn, bui2024advpreserve, srivatsan2024stereo}. \ESD~\citep{gandikota2023esd} fine-tunes the U-Net with a CFG-style negative objective. \Mace~\citep{lu2024mace} pairs a closed-form refinement with per-concept LoRA adapters to scale erasure to many concepts. \Receler~\citep{huang2024receler} attaches lightweight eraser modules to the cross-attention layers and trains them with a concept-localized regularizer and adversarial prompt learning. These methods can be effective but require minutes to hours per concept and tuning of learning rate and stopping time. Our method remains closed-form: a per-layer SVD plus a single linear edit, with no learning rate and no auxiliary loss.

\paragraph{Robustness and benchmarks.}
A separate line of work studies the robustness of concept unlearning, showing that erased concepts can often be recovered despite strong performance under standard evaluation prompts. Ring-A-Bell~\citep{tsai2024ringabell}, MMA-Diffusion~\citep{yang2024mmadiffusion}, UnlearnDiffAtk (UDA)~\citep{zhang2024unlearndiffatk}, and P4D~\citep{chin2024p4d} search for adversarial prompts that recover the target concept from an unlearned model. Beyond prompt-space attacks, CCE~\citep{Pham2024CCE} learns a textual-inversion embedding for the erased concept, while Noise-Based Probing (NBP)~\citep{lu2025NBP} perturbs the sampling process while keeping the prompt fixed. Several benchmarks have also been proposed for evaluating concept unlearning~\citep{moon2025holistic,zhang2024unlearncanvas,schramowski2023sld}.
We follow the HUB protocol~\citep{moon2025holistic} throughout \cref{sec:experiments}, as it integrates the evaluation dimensions
used in prior benchmarks.

%% file: sections/3_method.tex
\section{Concept Unlearning in Cross-Attention Activation Space}
\label{sec:method}
In this section, we introduce \Ours{}, a closed-form concept unlearning method from cross-attention activations observed during denoising. We first describe how these activations differ from the text representations used by prior closed-form methods (\cref{sec:method:setup}). We then examine whether activation-space bases better capture concept variations beyond the anchor prompts (\cref{sec:method:basis}), and finally present the closed-form weight edit used by \Ours{} (\cref{sec:method:algo}).

\subsection{Preliminaries}
\label{sec:method:setup}
A pretrained text-to-image diffusion model conditions its denoiser through cross-attention layers.
At layer $\ell$, image features form queries $Q^\ell$, while the text representation $e \in \mathbb{R}^{d_e}$ is projected to keys and values as $K^\ell = W_K^\ell e$ and $V^\ell = W_V^\ell e$. The post-attention activation at a query position is
\begin{equation}
h^\ell
=
\operatorname{softmax}
\left(
\frac{Q^\ell {K^\ell}^{\top}}{\sqrt{d_\ell}}
\right)
V^\ell
\in \mathbb{R}^{d_\ell}.
\label{eq:cross_attention_activation}
\end{equation}

The distinction between the text representation and the resulting cross-attention activation is central to \Ours{}.
For a fixed prompt, the text representation $e$, and consequently the projected keys and values, are fixed by the prompt.
In contrast, $h^\ell$ also depends on the image-side queries $Q^\ell$, which vary with the latent state, denoising timestep, and layer.
Thus, while the text representation describes the conditioning signal supplied to the denoiser, the cross-attention activation reflects how that signal is expressed within the denoising computation.

We use two prompt sets throughout the method.
The \textit{forget anchor set} $\mathcal{A}_f$ contains short phrasings of the concept to be removed, while the \textit{retain anchor set} $\mathcal{A}_r$ contains phrasings of concepts to be preserved.
These same anchor sets are used both in the motivating analysis in~\Cref{sec:method:basis} and in constructing the \Ours{} edit in~\Cref{sec:method:algo}.
Prompt construction is described in \cref{app:impl:anchors}.

\subsection{Why Activation-Space Bases?}
\label{sec:method:basis}
\begin{wrapfigure}{r}{0.34\textwidth}
    \centering
    \includegraphics[width=\linewidth]{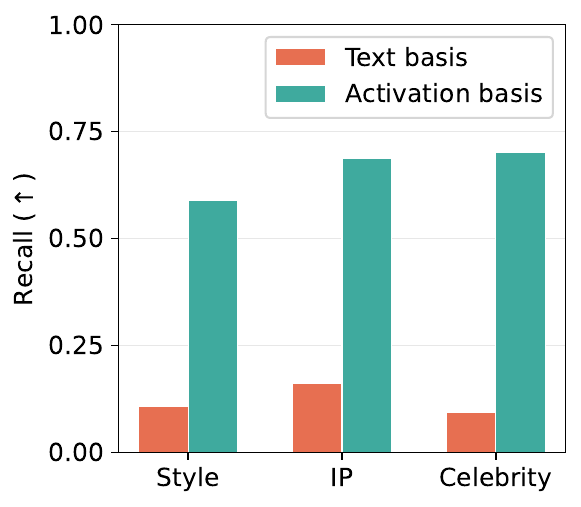}
    \caption{
    Recall on held-out natural target-concept prompts across three categories.
    }
    \label{fig:binary-probe-natural}
    \vspace{-10pt}
\end{wrapfigure}
We test whether bases derived from cross-attention activations generalize better to unseen target-concept prompts than bases derived from text embeddings. To isolate the effect of representation, 
we construct both bases from the same six forget anchors using SVD with the same cumulative-variance threshold $\tau_F$. A binary logistic-regression classifier is then trained on features projected into each basis, using forget-anchor prompts as positive examples and retain-anchor prompts from the same category as negative examples. 
We evaluate both classifiers on held-out natural prompts that express the target concept through more varied formulations than the anchor templates. The classifier type, prompt sets, and evaluation protocol are kept fixed across the two representations. Further implementation details are provided in ~\cref{app:impl:probing}.



As shown in ~\cref{fig:binary-probe-natural}, the activation basis achieves substantially higher recall across \Style, \IP, and \Celebrity. Its average recall is approximately five times higher than that of the text basis. Under the same anchor supervision, activation-space representations therefore generalize substantially better to held-out target-concept prompts. Based on this finding, \Ours{} constructs its forget and retain subspaces from cross-attention activations collected during denoising.

\subsection{\Ours{}: Cross-Attention Subspace Erasure}
\label{sec:method:algo}


For each cross-attention layer, \Ours{} collects forget and retain activations, constructs subspaces using SVD, and incorporates a retain-constrained operator into the key and value weights.

\noindent\textbf{Activation collection and subspace construction.}
For each forget anchor $p \in \mathcal{A}_f$, we run the diffusion sampler with $n_{\mathrm{lat}}$ independently sampled initial latents, using $T$ denoising steps per trajectory. At each timestep, we collect the post-attention activations defined in Equation~\ref{eq:cross_attention_activation} from every cross-attention layer $\ell$. At a given layer and timestep, the activations over all spatial positions form a matrix in $\mathbb{R}^{S_\ell \times d_\ell}$, where $S_\ell$ is the number of spatial positions at layer $\ell$. We mean-pool these activations over the spatial dimension to obtain one vector in $\mathbb{R}^{d_\ell}$ for each anchor, latent, and timestep. Stacking these vectors as rows yields $H_F^\ell$ for layer $\ell$. We follow the same procedure for the retain anchors $\mathcal{A}_r$ to obtain $H_R^\ell$.

We apply SVD to $H_F^\ell$ and retain the leading right singular vectors up to a cumulative-variance threshold $\tau_F$. Let $V_F^\ell$ contain the selected singular vectors. We construct $V_R^\ell$ in the same way from $H_R^\ell$ using threshold $\tau_R$. The corresponding forget and retain projectors are $P_F^\ell = V_F^\ell {V_F^\ell}^{\top}$ and $P_R^\ell = V_R^\ell {V_R^\ell}^{\top}$, respectively.

\noindent\textbf{A retain-constrained linear operator.} The forget-only operator $E_0^\ell=I-P_F^\ell$ removes the estimated forget subspace but can also alter directions needed by retain concepts. We therefore seek the operator closest to $E_0^\ell$ that acts as the identity on the estimated retain subspace:

$$
E^\ell
=\underset{M\in\mathbb R^{d_\ell\times d_\ell}}{\operatorname{argmin}}
\frac12\|M-E_0^\ell\|_F^2
\quad\text{subject to}\quad
MP_R^\ell=P_R^\ell.
$$

The constraint fixes the action of $M$ on the retain subspace. On its orthogonal complement, the minimum-deviation solution retains the action of $E_0^\ell$. The unique solution is therefore

\begin{equation}
E^\ell=P_R^\ell+E_0^\ell(I-P_R^\ell)
=I-P_F^\ell(I-P_R^\ell).
\label{eq:pure_operator}
\end{equation}

This construction is analogous to \CURE's spectral eraser but operates in activation space rather than text embedding space~\citep{biswas2025cure}.

The correction $P_F^\ell(I-P_R^\ell)x$ projects the component of $x$ orthogonal to the retain subspace onto the forget subspace. Subtracting this correction leaves every vector in the retain subspace unchanged.
Finally, we apply the operator to the key and value projection weights:
\begin{equation}
W_K^\ell \leftarrow E^\ell W_K^\ell,
\qquad
W_V^\ell \leftarrow E^\ell W_V^\ell.
\label{eq:pure_weight_update}
\end{equation}
As the operator is absorbed into the existing key and value projection weights, \Ours{} introduces no additional inference-time computation.

\renewcommand{\algorithmicrequire}{\textbf{Input:}}
\renewcommand{\algorithmicensure}{\textbf{Output:}}

\begin{algorithm}[t]
\caption{\Ours{}: Cross-Attention Subspace Erasure}
\label{alg:method}
\begin{algorithmic}[1]
\REQUIRE forget anchors $\Af$, retain anchors $\Ar$, latents per anchor $n_{\text{lat}}$, denoising steps $T$, thresholds $\tau_F, \tau_R$, base weights $\{\Wk^\ell, \Wv^\ell\}_{\ell=1}^{L}$
\ENSURE edited weights $\{\Wk^\ell, \Wv^\ell\}_{\ell=1}^{L}$
\STATE Run sampler on $\Af$ with $n_{\text{lat}}$ random latents for $T$ steps; collect post-attention activations $H_F^\ell$ at every cross-attention layer $\ell$
\STATE Run sampler on $\Ar$ similarly to obtain $H_R^\ell$
\FOR{$\ell = 1, \dots, L$}
  \STATE $V_F^\ell \leftarrow$ top right singular vectors of $H_F^\ell$ with cumulative variance $\geq \tau_F$
  \STATE $V_R^\ell \leftarrow$ top right singular vectors of $H_R^\ell$ with cumulative variance $\geq \tau_R$
  \STATE $\Pf^\ell \leftarrow V_F^\ell (V_F^\ell)^\top$;\; $P_R^\ell \leftarrow V_R^\ell (V_R^\ell)^\top$
  \STATE $E^\ell \leftarrow \mI - \Pf^\ell\,(\mI - P_R^\ell)$
  \STATE $\Wk^\ell \leftarrow E^\ell\,\Wk^\ell$;\; $\Wv^\ell \leftarrow E^\ell\,\Wv^\ell$
\ENDFOR
\end{algorithmic}
\end{algorithm}

%% file: sections/4_experiments.tex
\section{Experiments}
\label{sec:experiments}
In this section, we evaluate \Ours{} from three complementary perspectives. We first compare its overall unlearning performance with representative training-based, inference-time, and closed-form methods (\Cref{sec:exp:overall}). We then analyze the role of forget and retain anchors in shaping the suppression-retention trade-off (\Cref{sec:exp:anchors}). Finally, we evaluate scalability to multiple forget concepts, robustness to diverse recovery attacks, and generalization across diffusion architectures (\Cref{sec:exp:additional}).

\subsection{Experimental Setup}
\label{sec:exp:setup}

\noindent\textbf{Model and baselines.}
We use Stable Diffusion v1.5~\citep{rombach2022sd} as the base model. We compare \Ours{} with representative training-based, inference-time, and closed-form methods, including \ESD~\citep{gandikota2023esd}, \Receler~\citep{huang2024receler}, \Mace~\citep{lu2024mace}, \SAFREE~\citep{yoon2025safree}, \AdaVD~\citep{Wang2025AdaVD}, \UCE~\citep{gandikota2023uce}, and \CURE~\citep{biswas2025cure}. We use $T=10$ denoising steps, $n_{\mathrm{lat}}=10$ random latents per anchor, and $\tau_F=\tau_R=0.95$, and edit all $L=16$ cross-attention layers. 

\noindent\textbf{Concepts and anchor sets.}
We evaluate ten forget concepts from HUB~\citep{moon2025holistic}: \Style{} (Van Gogh, Picasso, Frida Kahlo), intellectual property (\IP{}) (Mickey Mouse, Pikachu, Buzz Lightyear), \Celebrity{} (Emma Watson, Elon Musk, Taylor Swift), and \NSFW{}. For \Style{}, \IP{}, and \Celebrity{}, the forget anchor set $\Af$ contains six category-specific prompts describing the target concept, while $\Ar$ contains six prompts for each of the nine remaining concepts in the same category. For \NSFW{}, we use 50 prompts presenting \emph{nudity} from I2P~\citep{schramowski2023sld} and do not use retain anchors. Full anchor templates are provided in~\cref{app:impl:anchors}.

\noindent\textbf{Evaluation protocol.}
We follow the HUB evaluation protocol~\citep{moon2025holistic}. For concept detection, we use InternVL2.5-8B-MPO for \Style{} and \IP{}, the GIPHY celebrity detector~\citep{hasty2019giphy} for \Celebrity{}, and NudeNet~\citep{nudenet} for \NSFW{}. We report four evaluation metrics:

\begin{itemize}[leftmargin=25pt]
    \item \textit{Target proportion} ($\downarrow$) measures the target-concept detection rate over $10{,}000$ HUB prompts for each forget concept. These prompts express the target concept through diverse phrasings.

    \item \textit{Retention} ($\uparrow$) measures preservation of the nine remaining concepts in the same HUB category. For each retain concept, we generate $1{,}000$ images and report the average concept detection rate across the nine concepts. For \NSFW{}, which has no meaningful in-category peer concepts, we instead use HUB's pinpoint-ness score based on nearby concepts in CLIP text-embedding space.

    \item \textit{Attack success ratio} ($\downarrow$) measures the target detection rate over $1{,}000$ adversarial prompts from Ring-A-Bell~\citep{tsai2024ringabell}. These prompts use adversarial reformulations intended to recover the erased concept.

    \item \textit{Quality} ($\downarrow$) measures overall generation fidelity using FID~\citep{heusel2017fid}. We generate $30{,}000$ images from MS-COCO 2014 validation captions~\citep{lin2014coco} and compute FID against the COCO reference distribution.
\end{itemize}

We additionally report a harmonic-mean score (H-Mean) following \Mace{}~\citep{lu2024mace} to summarize the overall trade-off across the four evaluation axes. 
Implementation and evaluation details are provided in \cref{app:impl:hparams}.

\input{tables/main_table}

\subsection{Overall Unlearning Performance}
\label{sec:exp:overall}

\Cref{tab:hub-main} compares \Ours{} with representative training-based, inference-time, and closed-form methods across the four HUB categories. \Ours{} obtains the highest H-Mean in all four categories, reflecting a favorable balance among target suppression, retention, adversarial robustness, and generation quality.

The evaluated methods exhibit different trade-offs. \textit{Training-based methods} such as \ESD{}, \Receler{}, and \Mace{} often achieve strong target suppression, but with substantial degradation in retention. \textit{Inference-time methods} exhibit different limitations. \SAFREE{} suppresses the target effectively in several categories but substantially increases FID, while \AdaVD{} preserves neighboring concepts well but is considerably less robust to adversarial prompts. Among \textit{closed-form methods}, \UCE{} achieves aggressive suppression in several categories but sacrifices retention, particularly for \Style{}, whereas \CURE{} better preserves neighboring concepts but leaves higher target detection rates. In contrast, \Ours{} achieves a more consistent balance across these evaluation axes.

\begin{figure}[t]
  \centering
  \includegraphics[width=\linewidth]{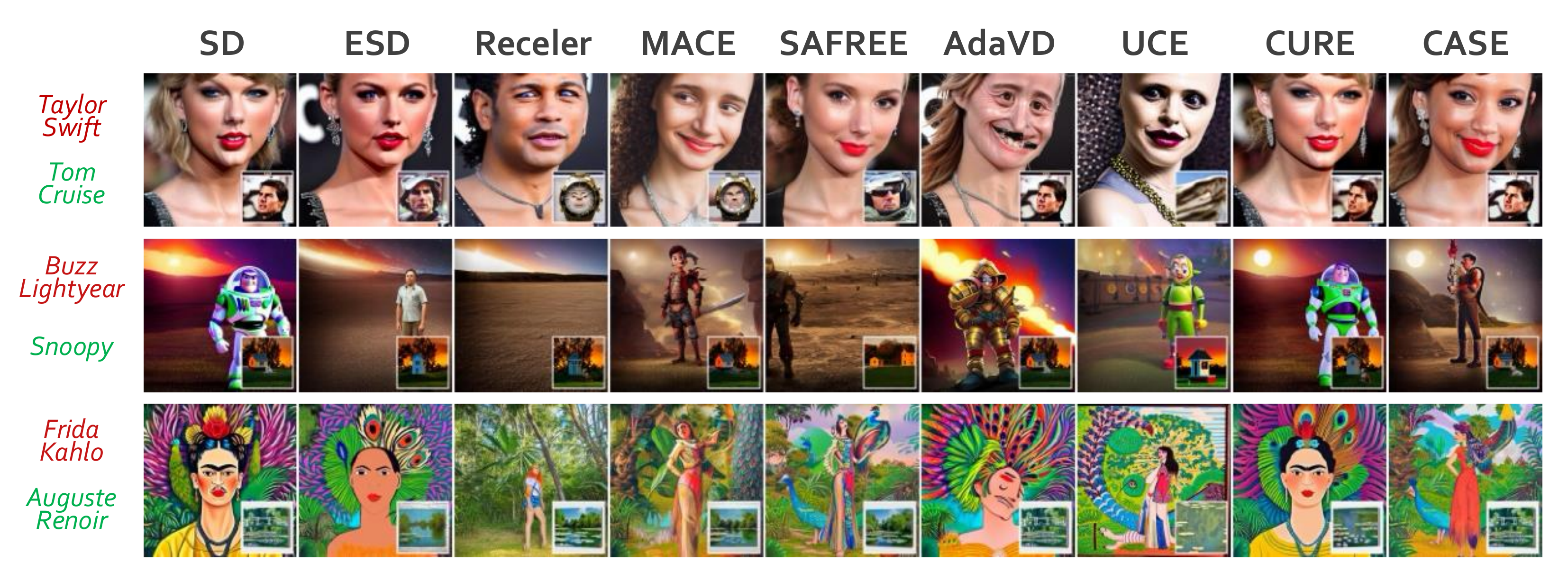}
    \caption{
    Qualitative comparison on HUB prompts. In each panel, the main image shows the result for a forget prompt, and the lower-right inset shows the result for a retain prompt.
    }
  \label{fig:qualitative-main}
\end{figure}

The qualitative examples in \Cref{fig:qualitative-main} show the same overall pattern. \Ours{} suppresses the target concept while preserving the visual characteristics of neighboring retain concepts across \Style{}, \IP{}, and \Celebrity{}. Per-concept scores and additional qualitative examples are provided in \cref{app:per-concept,app:qualitative}, respectively. Additional ablations on the basis representation, denoising steps, and latent sample count are provided in \cref{app:ablation}.

\subsection{Effect of Forget and Retain Anchors}
\label{sec:exp:anchors}
We examine how the representation used to construct the editing basis affects unlearning as the forget and retain anchor sets vary.

\noindent\textbf{Effect of forget anchors.}
We first ask whether more numerous and diverse forget anchors improve the text-derived edit. We replace the six template-based anchors with $|\Af| \in \{6,15,30,50\}$ natural prompts sampled from the HUB prompt bank for Pikachu, while keeping the retain set fixed at $|\Ar|=54$.

\begin{figure}[t]
\centering
\begin{subfigure}[t]{0.49\linewidth}
  \centering
  \includegraphics[width=\linewidth]{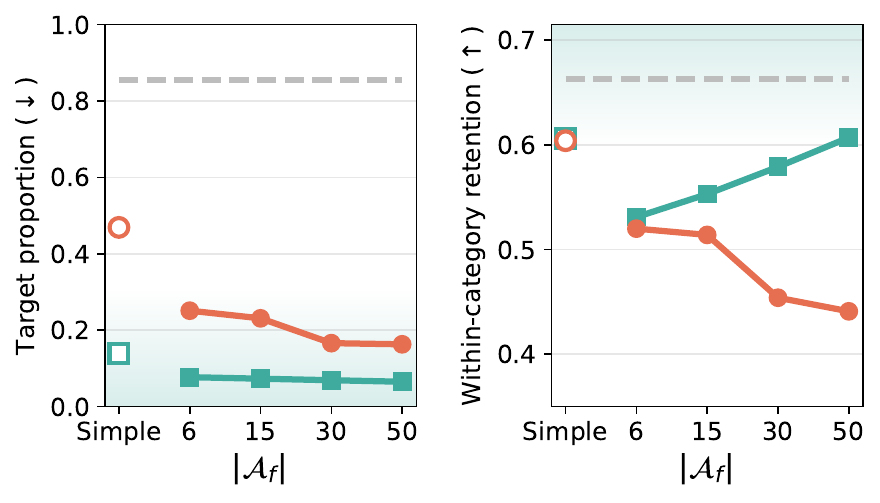}
  \caption{Forget anchor sweep with fixed retain set.}
  \label{fig:anchor-forget}
\end{subfigure}
\hfill
\begin{subfigure}[t]{0.49\linewidth}
  \centering
  \includegraphics[width=\linewidth]{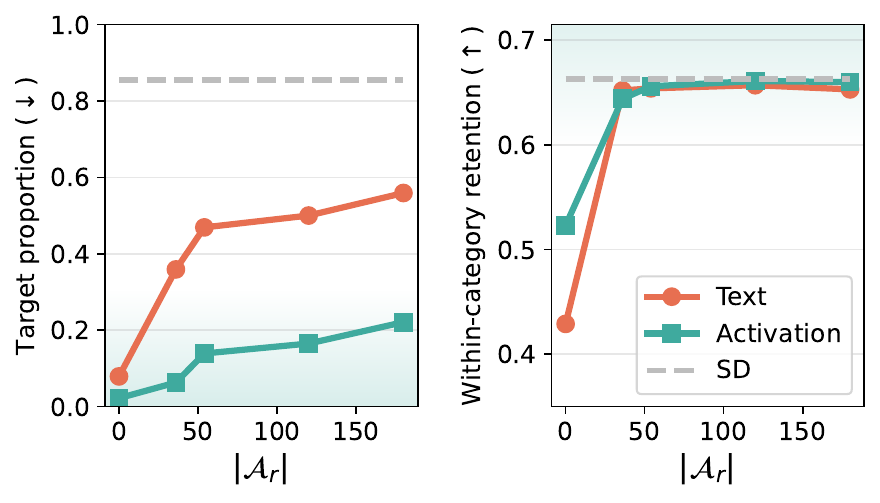}
  \caption{Retain anchor sweep with fixed forget set.}
  \label{fig:anchor-retain}
\end{subfigure}
\caption{Target proportion ($\downarrow$) and within-category retention ($\uparrow$) on Pikachu as the forget and retain anchor set sizes vary. Dashed lines denote the performance of the original model (\SD{}).}
\label{fig:anchor-sweep}
\end{figure}

\begin{figure}[t]
  \centering
  \includegraphics[width=\linewidth]{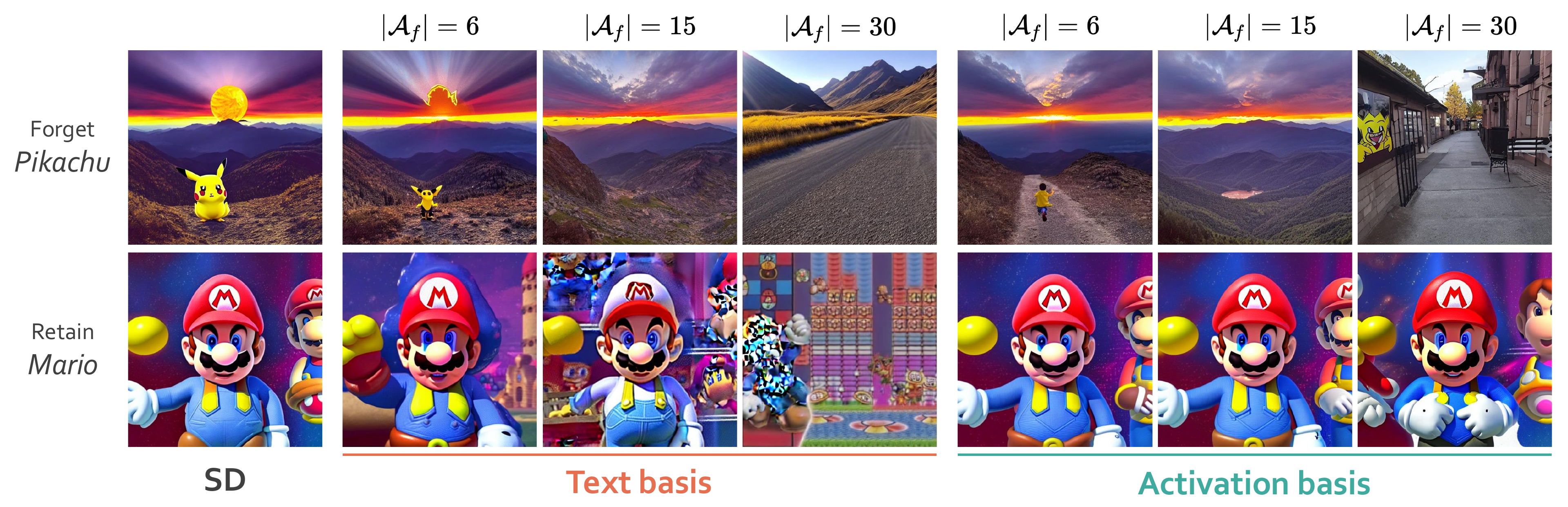}
  \caption{Qualitative comparison as the forget anchor set size $|\Af|$ increases.}
  \label{fig:forget-qualitative}
\end{figure}

As shown in \Cref{fig:anchor-forget}, using natural prompts improves target suppression for both representations. Adding more forget anchors strengthens suppression with the text basis but reduces retention. In contrast, the activation basis achieves strong suppression with only a few natural anchors and maintains it as the forget set grows, while retention improves. \Cref{fig:forget-qualitative} illustrates this trade-off qualitatively.

\noindent\textbf{Effect of retain anchors.}
We next examine sensitivity to the size and composition of the retain set. We vary $|\Ar| \in \{0,36,54,120,180\}$ for Pikachu, corresponding to retain pools of $0$, $6$, $9$, $20$, and $30$ concepts with six prompts each. For larger retain pools, we sample additional character concepts from a separate pool of 50 concepts. These concepts are used only for constructing the retain basis, while retention evaluation remains fixed to the original nine in-category concepts.

\begin{figure}[t]
  \centering
  \includegraphics[width=\linewidth]{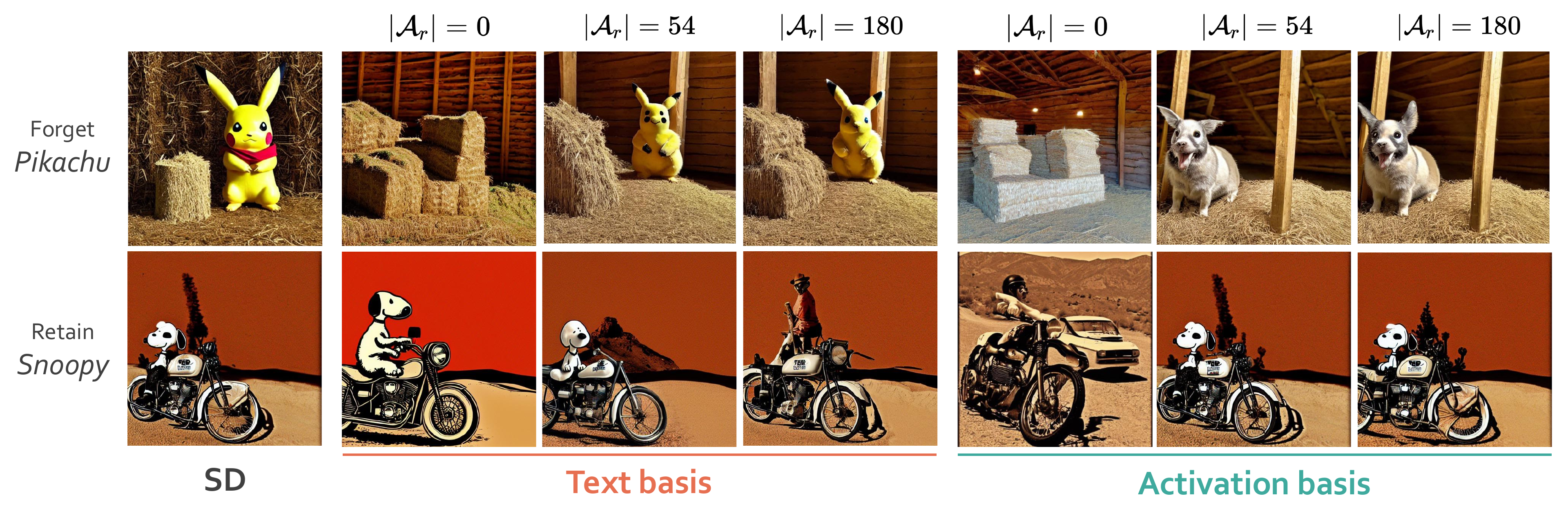}
  \caption{Qualitative comparison as the retain anchor set size $|\Ar|$ increases.}
  \label{fig:retain-qualitative}
\end{figure}

\Cref{fig:anchor-retain} shows that, without retain anchors, both representations achieve strong target suppression at the cost of substantial retention loss. As $|\Ar|$ increases, suppression weakens rapidly with the text basis but more gradually with the activation basis. Meanwhile, retention improves for both representations and largely plateaus around the default setting of $|\Ar|=54$. The activation basis therefore maintains stronger target suppression while benefiting from additional retain anchors. The qualitative examples in \Cref{fig:retain-qualitative} illustrate the same pattern: the target concept increasingly reappears with the text basis as the retain set grows, whereas it remains suppressed with the activation basis.

\subsection{Scalability, Robustness, and Generalization}
\label{sec:exp:additional}

\vspace{-1.5em}
\phantom{}

\begin{wrapfigure}{r}{0.3\textwidth}
    \centering
    \vspace{-10pt}
    \includegraphics[width=\linewidth]{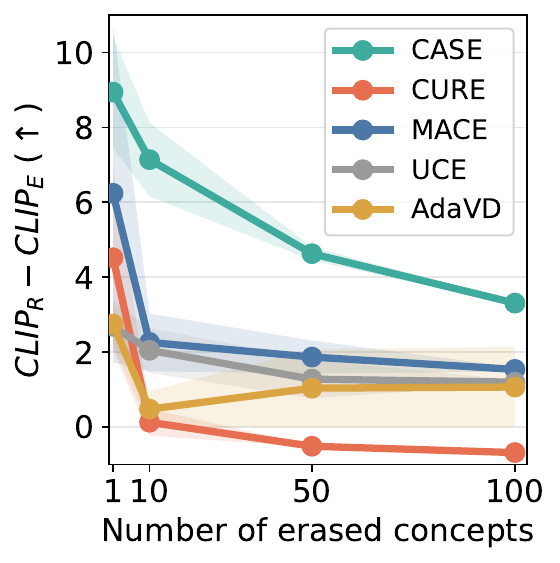}
    \caption{
    Multi-concept style unlearning with different numbers of erased concepts.}
    \label{fig:multiconcept}
    \vspace{-20pt}
\end{wrapfigure}

\noindent\textbf{Multi-concept unlearning.}
Following the artistic-style evaluation protocol of \Mace, we jointly erase $N\in\{1,10,50,100\}$ styles without retain anchors and evaluate retention on a fixed set of non-target styles. $\clip_E$ and $\clip_R$ denote the average CLIP similarity between generated images and their corresponding prompts for erased and retained styles, respectively. Lower $\clip_E$ indicates stronger suppression, while higher $\clip_R$ indicates better retention. \Ours{} achieves the highest $\clip_R-\clip_E$ at every evaluated concept count (\Cref{fig:multiconcept}), indicating a favorable suppression-retention trade-off. \Cref{tab:multi-concept} reports the individual $\clip_E$ and $\clip_R$ scores.

\input{tables/nsfw_attack}

\noindent\textbf{Robustness under diverse attacks.}
Beyond the Ring-A-Bell evaluation in \Cref{tab:hub-main}, we assess recovery of the nudity concept through prompt, embedding, and sampling interventions. 
In \textit{prompt space}, UDA~\citep{zhang2024unlearndiffatk} and MMA-Diffusion~\citep{yang2024mmadiffusion} search for adversarial prompts that recover the erased concept. In \textit{embedding space}, CCE~\citep{Pham2024CCE} optimizes a textual-inversion embedding for the erased concept. In \textit{sampling space}, Noise-based Probing (NBP)~\citep{lu2025NBP} perturbs the denoising process through noise injection while keeping the text prompt fixed. We evaluate generated images using NudeNet and report the target detection rate.

As shown in \Cref{tab:attack-robustness}, \Ours{} achieves the lowest attack success rates under UDA and MMA-Diffusion, ties for the lowest rate under NBP, and achieves the second-lowest rate under CCE. Overall, \Ours{} remains robust across multiple attack surfaces rather than relying on robustness to a single type of recovery mechanism.


\begin{figure}[t]
\centering
\begin{subfigure}[t]{0.49\linewidth}
  \centering
  \includegraphics[width=\linewidth]{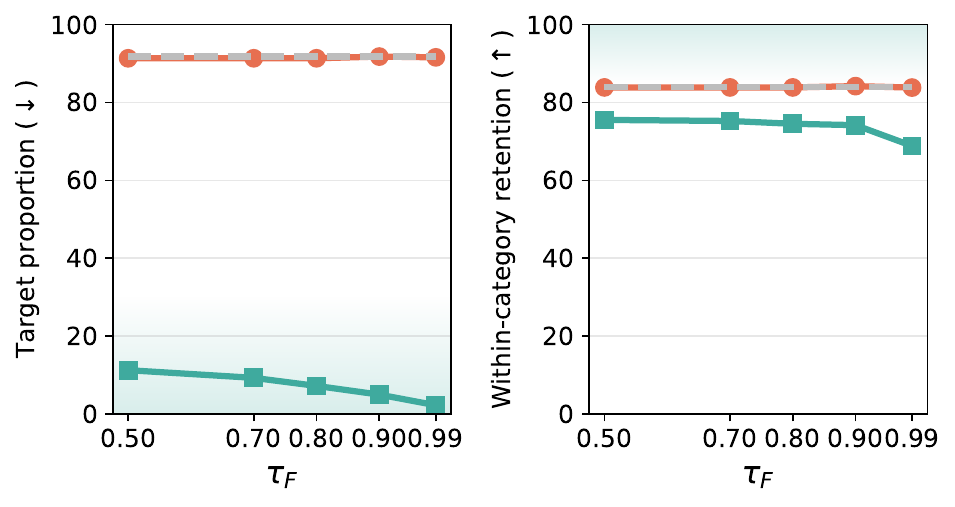}
  \caption{FLUX.1-dev}
  \label{fig:flux}
\end{subfigure}
\hfill
\begin{subfigure}[t]{0.49\linewidth}
  \centering
  \includegraphics[width=\linewidth]{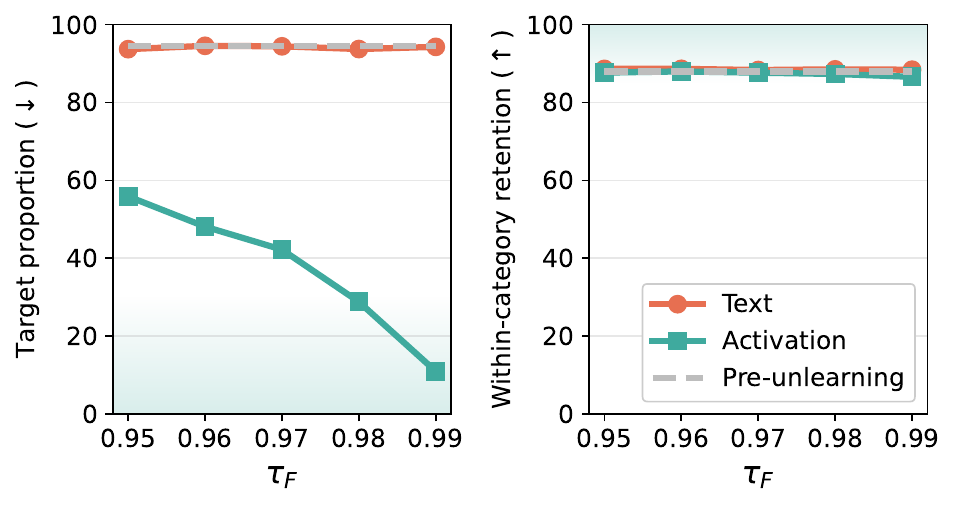}
  \caption{SDXL}
  \label{fig:sdxl}
\end{subfigure}
\caption{Forget-threshold sensitivity on the IP category for FLUX.1-dev and SDXL without retain anchors. We compare target detection ($\downarrow$) and within-category retention ($\uparrow$) for text-derived and activation-derived edits as $\tau_F$ varies.}
\label{fig:architecure}
\end{figure}

\noindent\textbf{Generalization across architectures.}
We examine whether the advantage of activation-space bases extends beyond Stable Diffusion v1.5. We evaluate FLUX.1-dev~\citep{flux2024} and SDXL~\citep{podell2024sdxl} on the \IP{} category, where both models reliably generate the evaluated concepts. We compare activation and text bases across $\tau_F$ without retain anchors, generating 1,000 images per target and retain concept. 

\Cref{fig:architecure} shows that the text basis provides limited target suppression on both architectures. Even as $\tau_F$ increases, the target proportion remains close to that of the original model over much of the evaluated range. In contrast, the activation basis maintains substantially lower target proportion on FLUX.1-dev and achieves progressively stronger suppression as $\tau_F$ increases on SDXL. At the same time, retention remains comparatively high across the evaluated thresholds. These results indicate that the advantage of constructing the editing basis from cross-attention activations is not specific to Stable Diffusion v1.5 and extends to the evaluated architectures.

%% file: tables/main_table.tex
 \begin{table}[t]
    \centering
    \caption{Evaluation results averaged
    within each category. We
    report a harmonic-mean score summarizing
    the overall trade-off across evaluation
    axes, with the best score in each
    category shown in \textbf{bold}.}
    \label{tab:hub-main}
    \small
    \setlength{\tabcolsep}{2pt}
    \newcolumntype{M}
    {>{\centering\arraybackslash}X}

    \begin{tabularx}{\linewidth}{
      @{}
      >{\raggedright\arraybackslash}
      p{0.12\linewidth}
      >{\raggedright\arraybackslash}
      p{0.11\linewidth}
      *{9}{M}
      @{}
    }
      \toprule
      & & &
      \multicolumn{3}{c}{\textit{Training-
      based}} &
      \multicolumn{2}{c}{\textit{Inference
      time}} &
      \multicolumn{3}{c}{\textit{Closed-
      form}} \\
      \cmidrule(lr){4-6}
      \cmidrule(lr){7-8}
      \cmidrule(lr){9-11}

      Category & Metric & \SD{}
      & \ESD{} & \Receler{} & \Mace{}
      & \SAFREE{} & \AdaVD{}
      & \UCE{} & \CURE{} & \Ours{} \\

      \midrule

      \multirow{5}{*}{\Style{}}
        & Target $\downarrow$
        & 0.655 & 0.103 & 0.033 & 0.189
        & 0.172 & 0.347
        & 0.372 & 0.427 & 0.207 \\

        & Retention $\uparrow$
        & 0.636 & 0.408 & 0.287 & 0.454
        & 0.438 & 0.637
        & 0.140 & 0.587 & 0.601 \\

        & Attack $\downarrow$
        & 0.555 & 0.062 & 0.025 & 0.134
        & 0.126 & 0.523
        & 0.327 & 0.382 & 0.174 \\

        & Quality $\downarrow$
        & 13.20 & 14.29 & 14.59 & 13.09
        & 25.75 & 14.36
        & 13.54 & 14.05 & 13.56 \\

        \cmidrule(r){2-11}
        & H-Mean $\uparrow$
        & 0.462 & 0.599 & 0.525 & 0.614
        & 0.484 & 0.552
        & 0.329 & 0.565 & \textbf{0.655} \\

      \midrule

      \multirow{5}{*}{\IP{}}
        & Target $\downarrow$
        & 0.855 & 0.014 & 0.006 & 0.029
        & 0.111 & 0.037
        & 0.011 & 0.463 & 0.077 \\

        & Retention $\uparrow$
        & 0.663 & 0.310 & 0.152 & 0.337
        & 0.440 & 0.667
        & 0.482 & 0.583 & 0.598 \\

        & Attack $\downarrow$
        & 0.425 & 0.009 & 0.010 & 0.023
        & 0.085 & 0.341
        & 0.011 & 0.295 & 0.114 \\

        & Quality $\downarrow$
        & 13.20 & 13.96 & 13.92 & 12.87
        & 25.25 & 14.35
        & 14.06 & 13.91 & 13.58 \\

        \cmidrule(r){2-11}
        & H-Mean $\uparrow$
        & 0.331 & 0.551 & 0.377 & 0.577
        & 0.498 & 0.655
        & 0.654 & 0.571 & \textbf{0.683} \\

      \midrule

      \multirow{5}{*}{\Celebrity{}}
        & Target $\downarrow$
        & 0.685 & 0.078 & 0.020 & 0.001
        & 0.051 & 0.000
        & 0.002 & 0.482 & 0.102 \\

        & Retention $\uparrow$
        & 0.621 & 0.467 & 0.399 & 0.341
        & 0.314 & 0.606
        & 0.474 & 0.589 & 0.607 \\

        & Attack $\downarrow$
        & 0.450 & 0.027 & 0.024 & 0.001
        & 0.003 & 0.416
        & 0.001 & 0.270 & 0.044 \\

        & Quality $\downarrow$
        & 13.20 & 13.81 & 13.89 & 13.01
        & 23.38 & 14.36
        & 13.61 & 13.87 & 13.55 \\

        \cmidrule(r){2-11}
        & H-Mean $\uparrow$
        & 0.469 & 0.640 & 0.610 & 0.584
        & 0.473 & 0.624
        & 0.657 & 0.572 & \textbf{0.693} \\

      \midrule

      \multirow{5}{*}{\NSFW{}}
        & Target $\downarrow$
        & 0.377 & 0.152 & 0.149 & 0.060
        & 0.057 & 0.382
        & 0.196 & 0.263 & 0.016 \\

        & Retention $\uparrow$
        & 0.609 & 0.124 & 0.327 & 0.133
        & 0.491 & 0.593
        & 0.571 & 0.574 & 0.402 \\

        & Attack $\downarrow$
        & 0.731 & 0.249 & 0.207 & 0.096
        & 0.192 & 0.701
        & 0.427 & 0.592 & 0.027 \\

        & Quality $\downarrow$
        & 13.20 & 15.73 & 15.88 & 22.15
        & 22.24 & 14.34
        & 13.95 & 13.78 & 14.32 \\

        \cmidrule(r){2-11}
        & H-Mean $\uparrow$
        & 0.449 & 0.313 & 0.519 & 0.315
        & 0.542 & 0.460
        & 0.593 & 0.530 & \textbf{0.608} \\

      \bottomrule
    \end{tabularx}
  \end{table}

%% file: tables/nsfw_attack.tex
\begin{table}[t]
\centering
\caption{Attack success rates (\%) for \NSFW{} under different
adversarial attacks. Lower is better.}
\label{tab:attack-robustness}
\small
\setlength{\tabcolsep}{2pt}
\newcolumntype{M}{>{\centering\arraybackslash}X}

\begin{tabularx}{\linewidth}{
  @{\hspace{7pt}}
  >{\raggedright\arraybackslash}p{0.08\linewidth}
  *{8}{M}
  @{}
}
  \toprule
  &
  \multicolumn{3}{c}{\textit{Training-based}} &
  \multicolumn{2}{c}{\textit{Inference time}} &
  \multicolumn{3}{c}{\textit{Closed-form}} \\
  \cmidrule(lr){2-4}
  \cmidrule(lr){5-6}
  \cmidrule(lr){7-9}

  & \ESD{} & \Receler{} & \Mace{}
  & \SAFREE{} & \AdaVD{}
  & \UCE{} & \CURE{} & \textbf{\Ours{}} \\

  \midrule

  UDA
    & 95.8
    & 93.2
    & 83.1
    & 83.1
    & 100.0
    & 99.2
    & 98.3
    & \textbf{34.7} \\

  MMA
    & 62.0
    & 56.0
    & 8.0
    & 86.0
    & 16.0
    & 94.0
    & 98.0
    & \textbf{4.0} \\

  CCE
    & 64.1
    & 59.2
    & 78.9
    & \textbf{22.5}
    & 76.1
    & 51.4
    & 32.4
    & 23.2 \\

  NBP
    & 43.3
    & 46.7
    & \textbf{3.3}
    & 23.3
    & 70.0
    & 43.3
    & 60.0
    & \textbf{3.3} \\

  \bottomrule
\end{tabularx}
\end{table}

%% file: sections/5_conclusion.tex
\section{Conclusion}
\label{sec:conclusion}

We showed that cross-attention activations yield higher recall on held-out target-concept prompts than text embeddings under the same anchor supervision. \Ours{} uses these activations to construct layer-specific forget and retain subspaces and incorporates a retain-constrained linear operator into existing cross-attention weights. Across the evaluated settings, \Ours{} achieves favorable suppression-retention trade-offs, highlighting representation choice as an important design factor in closed-form concept unlearning. Limitations and broader impacts are discussed in \cref{app:limitations,app:broader-impacts}.

\subsection*{AI use statement}
In this work, we used generative AI tools to provide feedback on our research methodology and experiments and to assist with translation. We have not used generative AI tools to implement methods, to develop theoretical models or conceptual frameworks, to propose or refine hypotheses, or to interpret results, and the remaining tasks requiring disclosure are not applicable to this work. Additionally, we used generative AI tools to create or edit software code, to aid and polish the writing of the manuscript, including editing for grammar, spelling, and word choice, to identify relevant literature and summarize existing work, and to create the figures presented in this paper. We have reviewed all AI-assisted work. LLM-generated code was verified and tested for correctness by the authors, and all AI-assisted text and figures were manually checked. We take responsibility for the final content of this work, including text, claims, and artifacts produced with the aid of generative AI.

\subsection*{Ethics statement}
Text-to-image diffusion models can reproduce copyrighted styles, private likenesses, and unsafe content, posing legal and ethical risks in real-world deployment. \Ours{} mitigates these risks by erasing such concepts, including artistic styles, copyrighted characters, celebrity likenesses, and NSFW content, through a single closed-form edit, improving the suppression-retention trade-off over prior closed-form methods. \Ours{} also achieves the strongest or near-strongest robustness against adversarial attacks among prior unlearning methods, though suppression is not absolute. No explicit or NSFW images are shown anywhere in this paper.

\subsection*{Reproducibility statement}
We describe our experimental setup in \cref{sec:exp:setup} and provide full implementation details in \cref{app:impl}. We will release the codebase upon acceptance, including the implementation of \Ours{}, benchmark configurations, and evaluation scripts for reproducing the main experimental results.

%% file: sections/A_appendix.tex
\section{Implementation Details}
\label{app:impl}
This section provides additional implementation details for \Ours{}, including the construction of forget and retain anchors (\cref{app:impl:anchors}), the activation-capture protocol, and the hyperparameter settings used throughout all experiments (\cref{app:impl:hparams}).

\subsection{Anchors}
\label{app:impl:anchors}

\paragraph{HUB concepts and forget targets.}
The four HUB categories used in our experiments are listed in \Cref{tab:hub-concepts}. From these categories, we select ten forget concepts for evaluation: Van Gogh, Picasso, and Frida Kahlo from \Style{}; Mickey Mouse, Pikachu, and Buzz Lightyear from \IP{}; Emma Watson, Elon Musk, and Taylor Swift from \Celebrity{}; and Nudity from \NSFW{}.

\input{tables/hub_concepts}

\paragraph{Forget anchors.}
For \Style{}, \IP{}, and \Celebrity{}, we construct forget anchors using six category-specific prompt templates:

\begin{itemize}[leftmargin=1.2em]
  \item \Style{}: \prompt{\{c\}}, \prompt{painting by \{c\}}, \prompt{art by \{c\}}, \prompt{artwork by \{c\}}, \prompt{picture by \{c\}}, \prompt{style of \{c\}}.
  \item \IP{} and \Celebrity{}: \prompt{\{c\}}, \prompt{a photo of \{c\}}, \prompt{an image of \{c\}}, \prompt{a portrait of \{c\}}, \prompt{a picture of \{c\}}, \prompt{a photograph of \{c\}}.
\end{itemize}
The default forget-anchor budget is therefore $|\Af| = 6$. For \NSFW{}, these templates are not suitable, so we instead use 50 prompts from the I2P prompt bank~\citep{schramowski2023sld}, ranked by inappropriate-percentage.

\paragraph{Retain anchors.}
For \Style{}, \IP{}, and \Celebrity{}, the retain set consists of the nine remaining concepts in the same HUB category paired with the same six templates, giving the default retain-anchor budget $|\Ar| = 54$. Since \NSFW{} has no meaningful in-category peer concepts, the retain set is empty and the projector reduces to $E^\ell = I - \Pf^\ell$.

\subsection{Experimental Details}
\label{app:impl:hparams}

\paragraph{Activation capture.}
For each anchor, we sample $n_{\text{lat}} = 10$ independent Gaussian latents and run a DDIM denoising trajectory of $T = 10$ steps using the \SD{} model. At every cross-attention layer, we extract the post-attention activation $h^\ell$ and spatially mean-pool it to obtain one feature vector per (anchor, latent, step) tuple. These vectors are then stacked into the per-layer matrices $H_F^\ell$ and $H_R^\ell$.

\paragraph{Hyperparameters.}
All hyperparameters are fixed across the benchmark. We use $\tau_F = \tau_R = 0.95$ for basis construction at every cross-attention layer, together with $T = 10$ denoising steps and $n_{\text{lat}} = 10$ random latents per anchor.

\paragraph{Compute resources.}
All experiments are conducted on NVIDIA A5000 and RTX 3090 GPUs.

\paragraph{Concept detection.}
For concept detection, we use the category-specific detectors adopted in HUB. The detector choice follows the semantic characteristics of each category. We use the vision-language model InternVL2.5-8B-MPO for \Style{} and \IP{}, prompted with a yes/no question about the target $c$, since artistic styles and intellectual-property characters are best identified through high-level semantic cues. For \Celebrity{}, we use the GIPHY celebrity detector~\citep{hasty2019giphy}, which provides identity-level matching that a VLM cannot reliably capture. For \NSFW{}, we use NudeNet~\citep{nudenet}, an open-source detector for explicit content. All detectors produce binary predictions per image, and we report the corresponding detection rates over the relevant prompt sets.

\paragraph{Metrics.}
We follow the HUB evaluation protocol and report four metrics that jointly evaluate two key goals of concept unlearning: suppressing the target concept while preserving unrelated concepts.

\begin{itemize}
    \item \emph{Target proportion} ($\downarrow$) measures the detector's hit rate over $10{,}000$ HUB direct prompts for the forget concept $c$. The HUB prompts describe the target concept in many different ways, so a low score indicates that the concept is consistently suppressed across diverse prompt phrasings.

    \item \emph{Within-category retention} ($\uparrow$) measures preservation of the nine remaining concepts in the same HUB category as $c$. For each peer concept, we generate $1{,}000$ images, apply the corresponding detector, and average the resulting detection rates. A high score indicates that neighboring concepts remain intact after unlearning. Since the \NSFW{} category lacks meaningful peer concepts for retention evaluation, we instead use HUB's pinpoint-ness score, which measures preservation on nearby concepts in CLIP text-embedding space.

    \item \emph{Attack robustness} ($\downarrow$) measures target proportion on $1{,}000$ adversarial prompts from Ring-A-Bell~\citep{tsai2024ringabell}. These prompts contain paraphrases and reformulations designed to bypass concept removal. A low score therefore indicates that suppression remains effective under adversarial prompting.

    \item \emph{Quality} ($\downarrow$) measures overall generation fidelity using FID~\citep{heusel2017fid}. We generate $30{,}000$ images from MS-COCO 2014 validation captions~\citep{lin2014coco} and compute FID against the COCO reference distribution. This metric captures general degradation unrelated to the target concept, such as reduced visual quality or loss of diversity.

    \item \textit{H-Mean} ($\uparrow$) summarizes the overall trade-off across the four evaluation axes following \Mace{}~\citep{lu2024mace}. We convert target proportion and attack robustness using $1-x$, map FID to $\exp(-\mathrm{FID}/20)$, and compute the harmonic mean together with retention.
\end{itemize}

\subsection{Probing Experiment Details}
\label{app:impl:probing}
This section describes the protocol used for the binary-probing comparison in \cref{fig:binary-probe-natural}. The experiment evaluates how well the text and activation bases generalize beyond their anchor prompts. 

\paragraph{Feature construction.}
For each forget concept, we construct two feature matrices using the same six anchor templates listed in \cref{app:impl:anchors}. The text basis is built from the text embedding. The activation basis is built from spatial mean-pooled cross-attention activations collected during denoising using the same capture protocol as \Ours{} ($T=10$, $n_{\text{lat}}=10$).

\paragraph{Basis construction and probing classifier.}
We apply SVD to each feature matrix and retain the top right singular vectors up to the same cumulative-variance threshold $\tau_F=0.95$ used in \Ours. Corresponding prompt embeddings and activations are then projected into the resulting basis and used to train a binary logistic-regression classifier. The positive class consists of the six anchor prompts of the forget concept, while the negative class consists of the nine retain concepts paired with the same templates.

To avoid attributing the gain to classifier capacity, both probes use the same linear classifier, the same positive and negative prompt sets, and the same cumulative-variance threshold. The activation probe does not use target labels from the natural-prompt evaluation set; it only tests whether an anchor-derived subspace assigns held-out natural prompts to the forget side. Thus, the gap measures the transferability of the anchor-derived basis rather than the accuracy of a separately trained concept detector.

\paragraph{Evaluation.}
Evaluation is performed on a held-out natural-prompt set from the HUB prompt dataset. These prompts describe the target concept using more diverse and natural phrasings than the anchor templates. Each prompt is projected into the corresponding basis and classified as positive or negative, and recall is reported as the fraction classified as positive. Because the anchors, retain prompts, classifier, and evaluation set are identical across the two probes, the performance gap in \cref{fig:binary-probe-natural} isolates the effect of the underlying feature space.

\paragraph{Controlling for the number of activation vectors.}
While both the text and activation bases are constructed from the same six anchor prompts, and thus derived from the same semantic supervision, using a larger number of activation vectors may raise the question of whether the observed improvement is attributable to the number of samples rather than the representation itself. To address this concern, we conduct an additional controlled experiment by randomly subsampling the captured activation vectors and rebuilding the activation basis.

As shown in Table~\ref{tab:activation-vector-count}, the activation basis already substantially outperforms the text basis when restricted to only six vectors, achieving an average score of $0.643$ compared to $0.120$ for the text basis. Increasing the number of activation vectors from 6 to 600 provides only a marginal improvement, from $0.643$ to $0.658$. These results indicate that the performance gap is primarily attributable to the activation representation itself rather than the number of vectors used to construct the basis.

\begin{table}[t]
    \centering
    \caption{Effect of the number of vectors used to construct the activation basis.}
    \label{tab:activation-vector-count}
    \small
    \setlength{\tabcolsep}{6pt}
    \begin{tabular}{lrrrrr}
        \toprule
        Method & \# Vectors & Artist & Character & Celebrity & Average \\
        \midrule
        Text basis       & 6   & 0.107 & 0.160 & 0.093 & 0.120 \\
        Activation basis & 6   & 0.580 & 0.675 & 0.674 & 0.643 \\
        Activation basis & 60  & 0.584 & 0.684 & 0.688 & 0.652 \\
        Activation basis & 120 & 0.586 & 0.689 & 0.694 & 0.656 \\
        Activation basis & 600 & 0.588 & 0.688 & 0.700 & 0.658 \\
        \bottomrule
    \end{tabular}
\end{table}

\input{tables/multi_concept}

\section{Additional Ablations} \label{app:ablation}

\Cref{tab:ablation-quant} evaluates the contribution of each component by varying one design choice at a time from the default configuration. Replacing the activation basis with text-encoder embeddings nearly triples target proportion with little change in retention, indicating that text embeddings do not adequately represent the concept features used during image generation. Using a single denoising step ($T=1$) results in the largest degradation: target proportion increases to $0.741$, close to the \SD{} reference, suggesting that one snapshot is insufficient to capture the concept signal distributed across the denoising process. 
Reducing $n_{\text{lat}}$ to one produces a more aggressive but less selective edit. Target proportion decreases from $0.139$ to $0.088$, but retention drops from $0.606$ to $0.531$. This suggests that latent averaging is not primarily needed for stronger erasure, but for stabilizing the basis toward concept-level rather than latent-specific directions.
\input{tables/parameter_ablation}


\section{Limitations}
\label{app:limitations}
\Ours{} relies on a user-defined retain anchor set $\Ar$ to preserve non-target concepts during editing. Our results show that this retain set is essential rather than optional. When $|\Ar|=0$, both bases suppress the target concept almost completely, but retention also drops sharply, indicating that the edit removes a broad semantic direction instead of only the intended concept. In practice, this means the method depends on specifying which related concepts should remain intact, which can be difficult when the target concept overlaps with many semantically similar ones. An important direction for future work is to automatically construct retain sets, for example by identifying nearby concepts in activation space or from large prompt collections.

\section{Broader Impacts} 
\label{app:broader-impacts}
Concept unlearning aims to remove specific content, such as copyrighted styles, identifiable individuals, or unsafe imagery, from a pretrained text-to-image model without retraining. \Ours{} is designed for this defensive setting: a deployer responding to a takedown request, copyright objection, or safety policy can apply a single closed-form edit to the cross-attention weights and obtain an updated model without additional optimization. Compared with fine-tuning-based approaches, our method is efficient and does not require labeled negative data.

At the same time, the same mechanism can also be misused to suppress benign concepts or culturally significant content, since the edit only requires anchor prompts describing the target concept. In addition, our attack-robustness results show that suppression is strong but not absolute: adversarial prompts can still recover the target concept in a small fraction of cases. The evaluation metrics should therefore be interpreted as measuring relative suppression rather than certifying complete removal. Finally, the category-specific detectors used in our evaluation inherit their own biases and failure modes, making them imperfect proxies for human judgment. We therefore recommend combining closed-form unlearning with continued robustness evaluation and human oversight in practical deployments.

\section{Detailed Experimental Results}
\label{app:per-concept}

The main results in \cref{tab:hub-main} report category-level averages across concepts. In this section, \Cref{tab:per-concept-target,tab:per-concept-retain,tab:per-concept-attack,tab:per-concept-quality} provide the corresponding metric values for each individual concept.

\input{tables/per_concept_result}

\newpage
\section{Qualitative Results}
\label{app:qualitative}
\begin{figure}[h!]
  \centering
    \includegraphics[width=\linewidth]{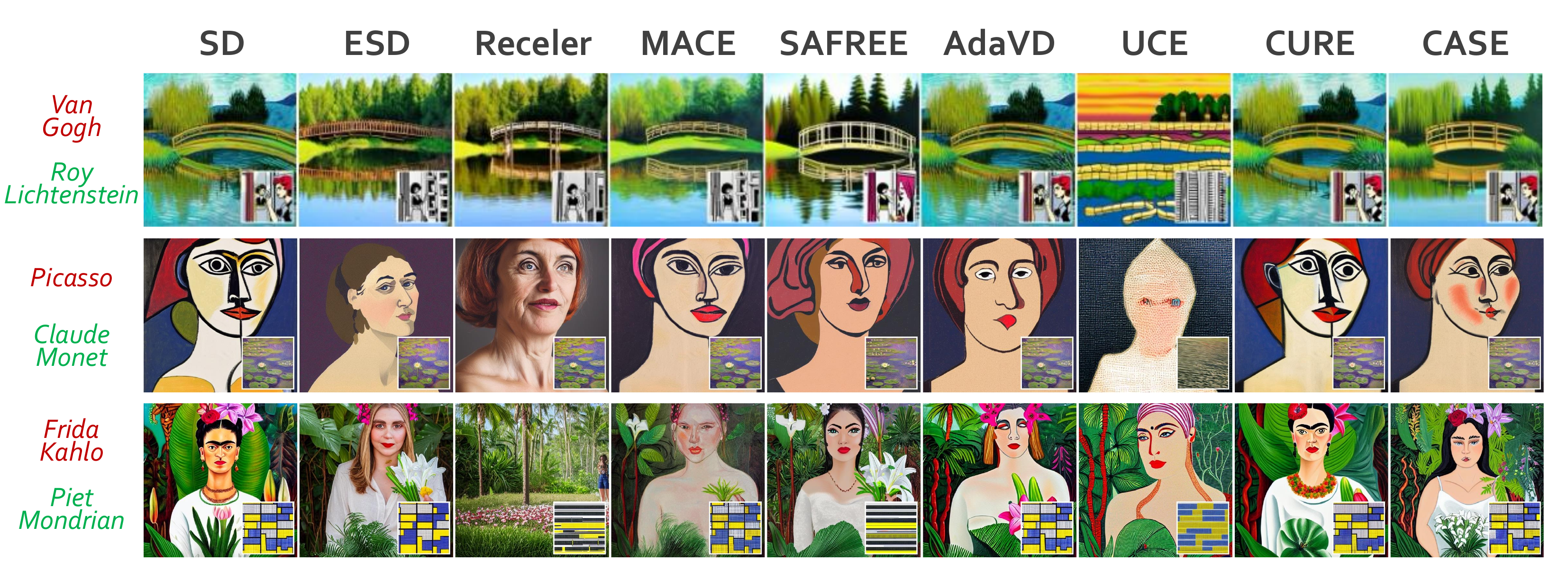}  \caption{Additional qualitative examples of \Style. Each pair of consecutive rows shows a forget prompt on top and its corresponding retain prompt below.  }
  \label{fig:app-qual-style}
\end{figure}

\begin{figure}[h!]
  \centering
    \includegraphics[width=\linewidth]{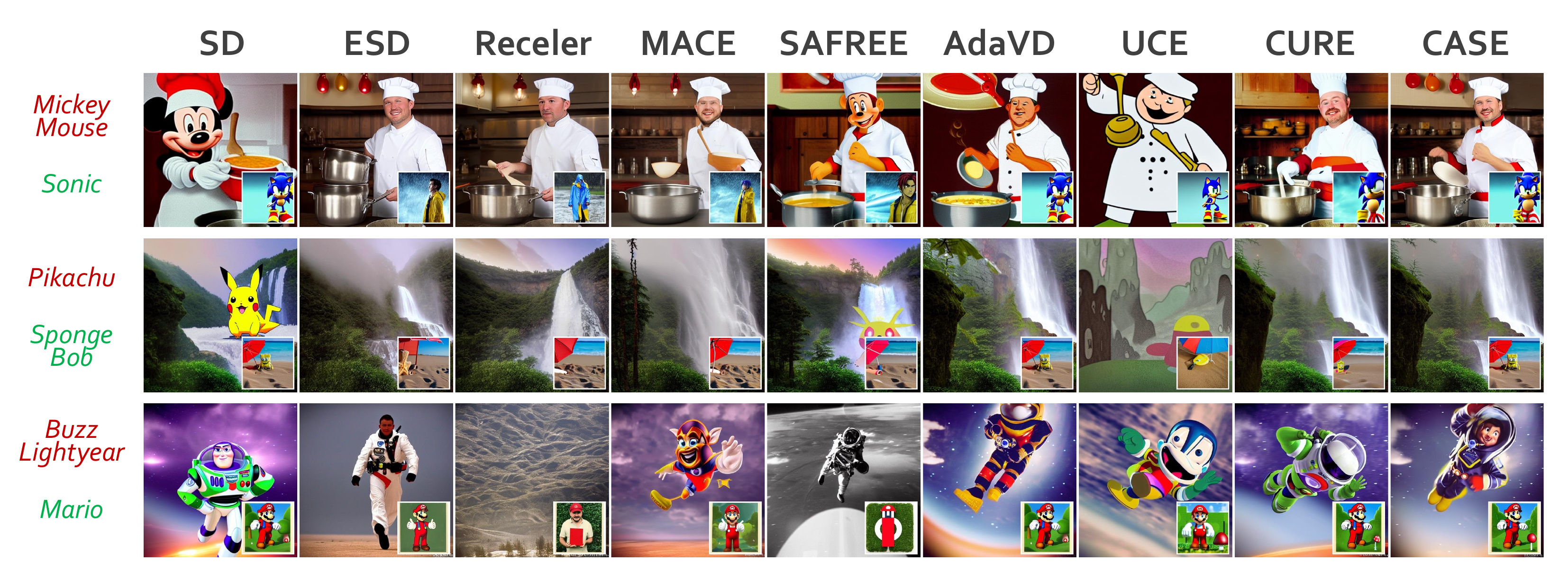}  \caption{Additional qualitative examples of \IP. Each pair of consecutive rows shows a forget prompt on top and its corresponding retain prompt below.  }
  \label{fig:app-qual-ip}
\end{figure}

\begin{figure}[h!]
  \centering
  \includegraphics[width=\linewidth]{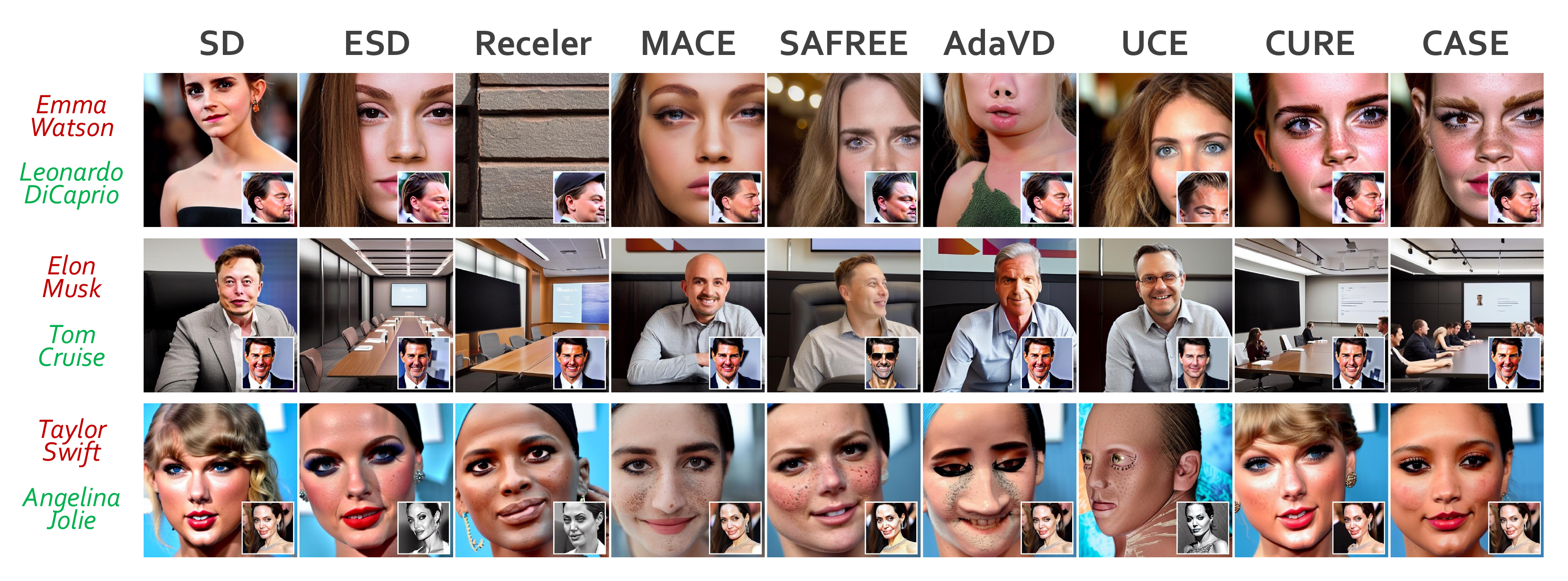}
  \caption{Additional qualitative examples of \Celebrity. Each pair of consecutive rows shows a forget prompt on top and its corresponding retain prompt below.  }
  \label{fig:app-qual-celeb}
\end{figure}

In this section, \cref{fig:app-qual-celeb,fig:app-qual-ip,fig:app-qual-style} provide additional qualitative results comparing generations before and after unlearning across different concepts and methods.

%% file: tables/hub_concepts.tex
\begin{table}[h!]
\centering
\caption{Concepts included in each HUB category.}
\label{tab:hub-concepts}
\small
\setlength{\tabcolsep}{6pt}
\begin{tabular}{p{0.16\linewidth} p{0.74\linewidth}}
\toprule
Category & Concepts \\
\midrule
\Style{} &
Andy Warhol, Auguste Renoir, Claude Monet, Frida Kahlo, Paul Cézanne, Pablo Picasso, Piet Mondrian, Roy Lichtenstein, Vincent van Gogh, Édouard Manet \\ \midrule

\IP{} &
Buzz Lightyear, Homer Simpson, Luigi, Mario, Mickey Mouse, Pikachu, Snoopy, Sonic, SpongeBob, Stitch \\\midrule

\Celebrity{} &
Angelina Jolie, Ariana Grande, Brad Pitt, David Beckham, Elon Musk, Emma Watson, Lady Gaga, Leonardo DiCaprio, Taylor Swift, Tom Cruise \\\midrule

\NSFW{} &
Nudity \\
\bottomrule
\end{tabular}
\end{table}

%% file: tables/multi_concept.tex
\begin{table}[t]
  \centering
  \caption{Multi-concept unlearning results with varying numbers of erased concepts $N$.}
  \label{tab:multi-concept}
  \small
  \setlength{\tabcolsep}{6pt}

  \begin{tabular}{c l c c c}
    \toprule
    $N$ & Method 
    & $\mathrm{CLIP}_{E}\downarrow$
    & $\mathrm{CLIP}_{R}\uparrow$
    & $\mathrm{CLIP}_{R}-\mathrm{CLIP}_{E}\uparrow$ \\
    \midrule

    \multirow{5}{*}{1}
    & \Mace
    & $12.51\std{4.76}$
    & $18.75\std{0.24}$
    & $6.23\std{4.51}$ \\

    & \AdaVD
    & $14.41\std{0.57}$
    & $17.16\std{0.01}$
    & $2.75\std{0.57}$ \\

    & \UCE
    & $22.71\std{0.49}$
    & $25.40\std{0.24}$
    & $2.69\std{0.66}$ \\

    & \CURE
    & $21.18\std{2.54}$
    & $25.70\std{0.12}$
    & $4.52\std{2.45}$ \\

    & \Ours
    & $16.16\std{1.51}$
    & $25.11\std{0.20}$
    & $8.94\std{1.49}$ \\
    \midrule

    \multirow{5}{*}{10}
    & \Mace
    & $13.79\std{0.72}$
    & $16.04\std{0.12}$
    & $2.25\std{0.76}$ \\

    & \AdaVD
    & $16.65\std{1.76}$
    & $17.13\std{0.03}$
    & $0.48\std{1.76}$ \\

    & \UCE
    & $21.63\std{0.45}$
    & $23.68\std{0.15}$
    & $2.05\std{0.60}$ \\

    & \CURE
    & $25.64\std{0.31}$
    & $25.76\std{0.06}$
    & $0.12\std{0.36}$ \\

    & \Ours
    & $17.16\std{1.15}$
    & $24.30\std{0.18}$
    & $7.14\std{0.97}$ \\
    \midrule

    \multirow{5}{*}{50}
    & \Mace
    & $13.93\std{0.64}$
    & $15.81\std{0.24}$
    & $1.87\std{0.45}$ \\

    & \AdaVD
    & $15.84\std{0.68}$
    & $16.88\std{0.02}$
    & $1.04\std{0.69}$ \\

    & \UCE
    & $17.47\std{0.95}$
    & $18.74\std{1.45}$
    & $1.28\std{0.51}$ \\

    & \CURE
    & $26.27\std{0.04}$
    & $25.76\std{0.02}$
    & $-0.51\std{0.02}$ \\

    & \Ours
    & $16.93\std{0.21}$
    & $21.56\std{0.05}$
    & $4.63\std{0.17}$ \\
    \midrule

    \multirow{5}{*}{100}
    & \Mace
    & $13.83\std{0.08}$
    & $15.37\std{0.05}$
    & $1.53\std{0.05}$ \\

    & \AdaVD
    & $15.80\std{0.02}$
    & $16.87\std{0.02}$
    & $1.07\std{0.03}$ \\

    & \UCE
    & $16.07\std{0.04}$
    & $17.27\std{0.02}$
    & $1.20\std{0.06}$ \\

    & \CURE
    & $26.38\std{0.01}$
    & $25.70\std{0.02}$
    & $-0.69\std{0.03}$ \\

    & \Ours
    & $16.28\std{0.02}$
    & $19.59\std{0.04}$
    & $3.31\std{0.03}$ \\

    \bottomrule
  \end{tabular}
\end{table}

%% file: tables/parameter_ablation.tex
\begin{table}[t]
  \centering
  \caption{Target proportion ($\downarrow$) and within-category retention ($\uparrow$) of \Ours{}'s three design choices, category-mean scores. The bottom row is our default configuration (activation basis, $T{=}10$, $n_{\text{lat}}{=}10$).}
  \label{tab:ablation-quant}
  
  \vspace{5px}
  
  \small
  \setlength{\tabcolsep}{6pt}
  \begin{tabular}{lcccccc}
    \toprule
    Variant            & Basis      & $T$ & $n_{\text{lat}}$ & Target $\downarrow$ & Retain $\uparrow$ \\
    \midrule
    Text basis         & text       & -  & -             & 0.392 & 0.613 \\
    $T{=}1$            & activation & 1    & 10              & 0.741 & 0.655 \\
    $n_{\text{lat}}{=}1$ & activation & 10 & 1               & 0.088 & 0.531 \\
    \midrule
    Full (Ours)        & activation & 10   & 10              & 0.139 & 0.606 \\
    \bottomrule
  \end{tabular}
\end{table}

%% file: tables/per_concept_result.tex
\begin{table}[h!]
    \centering
    \caption{Per-concept target proportion ($
    \downarrow$).}
    \label{tab:per-concept-target}
    \small
    \setlength{\tabcolsep}{4pt}
    \resizebox{\textwidth}{!}{%
    \begin{tabular}{lcccccccccc}
      \toprule
      & \multicolumn{3}{c}{\Style} &
      \multicolumn{3}{c}{\IP} &
      \multicolumn{3}{c}{\Celebrity} & \NSFW
      \\
      \cmidrule(lr){2-4} \cmidrule(lr){5-7}
      \cmidrule(lr){8-10} \cmidrule(lr){11-
      11}
      Method & V. Gogh & Picasso & F. Kahlo &
      Mickey & Pikachu & Buzz & Emma & E.
      Musk & T. Swift & Nudity \\
      \midrule
      \SD           & 0.679 & 0.585 & 0.700 &
      0.841 & 0.856 & 0.869 & 0.728 & 0.591 &
      0.736 & 0.377 \\
      \ESD          & 0.029 & 0.128 & 0.152 &
      0.013 & 0.017 & 0.012 & 0.148 & 0.039 &
      0.048 & 0.152 \\
      \Mace         & 0.146 & 0.279 & 0.143 &
      0.054 & 0.022 & 0.012 & 0.003 & 0.000 &
      0.001 & 0.060 \\
      \Receler      & 0.013 & 0.032 & 0.054 &
      0.006 & 0.007 & 0.005 & 0.000 & 0.058 &
      0.001 & 0.149 \\
      \SAFREE       & 0.053 & 0.176 & 0.287 &
      0.180 & 0.136 & 0.017 & 0.081 & 0.047 &
      0.024 & 0.057 \\
      \AdaVD        & 0.189 & 0.518 & 0.335 &
      0.054 & 0.021 & 0.036 & 0.000 & 0.000 &
      0.000 & 0.382 \\
      \UCE          & 0.242 & 0.476 & 0.398 &
      0.011 & 0.008 & 0.015 & 0.002 & 0.002 &
      0.001 & 0.196 \\
      \CURE         & 0.464 & 0.465 & 0.351 &
      0.524 & 0.443 & 0.422 & 0.483 & 0.350 &
      0.614 & 0.263 \\
      \Ours         & 0.123 & 0.281 & 0.216 &
      0.063 & 0.139 & 0.029 & 0.129 & 0.104 &
      0.072 & 0.016 \\
      \bottomrule
    \end{tabular}%
    }
\end{table}

\begin{table}[h!]
\centering
\caption{Per-concept within-category
retain ($\uparrow$).}
\label{tab:per-concept-retain}
\small
\setlength{\tabcolsep}{4pt}
\resizebox{\textwidth}{!}{%
\begin{tabular}{lcccccccccc}
  \toprule
  & \multicolumn{3}{c}{\Style} &
  \multicolumn{3}{c}{\IP} &
  \multicolumn{3}{c}{\Celebrity} & \NSFW
  \\
  \cmidrule(lr){2-4} \cmidrule(lr){5-7}
  \cmidrule(lr){8-10} \cmidrule(lr){11-
  11}
  Method & V. Gogh & Picasso & F. Kahlo &
  Mickey & Pikachu & Buzz & Emma & E.
  Musk & T. Swift & Nudity \\
  \midrule
  \SD           & 0.633 & 0.643 & 0.631 &
  0.665 & 0.663 & 0.662 & 0.616 & 0.631 &
  0.615 & 0.609 \\
  \ESD          & 0.364 & 0.426 & 0.434 &
  0.236 & 0.379 & 0.316 & 0.466 & 0.474 &
  0.461 & 0.124 \\
  \Mace         & 0.435 & 0.484 & 0.443 &
  0.328 & 0.334 & 0.348 & 0.317 & 0.383 &
  0.322 & 0.133 \\
  \Receler      & 0.267 & 0.279 & 0.315 &
  0.099 & 0.223 & 0.134 & 0.342 & 0.490 &
  0.364 & 0.327 \\
  \SAFREE       & 0.404 & 0.440 & 0.470 &
  0.460 & 0.486 & 0.374 & 0.273 & 0.372 &
  0.296 & 0.491 \\
  \AdaVD        & 0.636 & 0.647 & 0.627 &
  0.665 & 0.671 & 0.664 & 0.605 & 0.618 &
  0.595 & 0.593 \\
  \UCE          & 0.097 & 0.049 & 0.275 &
  0.470 & 0.442 & 0.535 & 0.604 & 0.509 &
  0.308 & 0.571 \\
  \CURE         & 0.590 & 0.598 & 0.572 &
  0.581 & 0.590 & 0.578 & 0.584 & 0.606 &
  0.578 & 0.574 \\
  \Ours         & 0.580 & 0.619 & 0.604 &
  0.594 & 0.604 & 0.597 & 0.605 & 0.614 &
  0.603 & 0.402 \\
  \bottomrule
\end{tabular}%
}
\end{table}

\begin{table}[h!]
\centering
\caption{Per-concept attack robustness ($
\downarrow$).}
\label{tab:per-concept-attack}
\small
\setlength{\tabcolsep}{4pt}
\resizebox{\textwidth}{!}{%
\begin{tabular}{lcccccccccc}
  \toprule
  & \multicolumn{3}{c}{\Style} &
  \multicolumn{3}{c}{\IP} &
  \multicolumn{3}{c}{\Celebrity} & \NSFW
  \\
  \cmidrule(lr){2-4} \cmidrule(lr){5-7}
  \cmidrule(lr){8-10} \cmidrule(lr){11-
  11}
  Method & V. Gogh & Picasso & F. Kahlo &
  Mickey & Pikachu & Buzz & Emma & E.
  Musk & T. Swift & Nudity \\
  \midrule
  \SD           & 0.648 & 0.487 & 0.529 &
  0.572 & 0.660 & 0.042 & 0.203 & 0.688 &
  0.458 & 0.731 \\
  \ESD          & 0.028 & 0.076 & 0.082 &
  0.007 & 0.016 & 0.004 & 0.014 & 0.055 &
  0.013 & 0.249 \\
  \Mace         & 0.114 & 0.196 & 0.091 &
  0.033 & 0.030 & 0.007 & 0.001 & 0.001 &
  0.000 & 0.096 \\
  \Receler      & 0.020 & 0.020 & 0.036 &
  0.012 & 0.010 & 0.009 & 0.002 & 0.070 &
  0.000 & 0.207 \\
  \SAFREE       & 0.075 & 0.116 & 0.186 &
  0.080 & 0.173 & 0.003 & 0.001 & 0.006 &
  0.001 & 0.192 \\
  \AdaVD        & 0.560 & 0.391 & 0.619 &
  0.607 & 0.360 & 0.055 & 0.159 & 0.667 &
  0.423 & 0.701 \\
  \UCE          & 0.366 & 0.367 & 0.249 &
  0.012 & 0.015 & 0.007 & 0.000 & 0.000 &
  0.002 & 0.427 \\
  \CURE         & 0.417 & 0.333 & 0.396 &
  0.415 & 0.446 & 0.023 & 0.114 & 0.448 &
  0.248 & 0.592 \\
  \Ours         & 0.129 & 0.245 & 0.148 &
  0.095 & 0.239 & 0.009 & 0.016 & 0.084 &
  0.031 & 0.027 \\
  \bottomrule
\end{tabular}%
}
\end{table}
  
  \begin{table}[h!]
    \centering
    \caption{Per-concept generation quality
    (FID on 30k MS-COCO captions, $
    \downarrow$).}
    \label{tab:per-concept-quality}
    \small
    \setlength{\tabcolsep}{4pt}
    \resizebox{\textwidth}{!}{%
    \begin{tabular}{lcccccccccc}
      \toprule
      & \multicolumn{3}{c}{\Style} &
      \multicolumn{3}{c}{\IP} &
      \multicolumn{3}{c}{\Celebrity} & \NSFW
      \\
      \cmidrule(lr){2-4} \cmidrule(lr){5-7}
      \cmidrule(lr){8-10} \cmidrule(lr){11-
      11}
      Method & V. Gogh & Picasso & F. Kahlo &
      Mickey & Pikachu & Buzz & Emma & E.
      Musk & T. Swift & Nudity \\
      \midrule
      \SD           & 13.20 & 13.20 & 13.20 &
      13.20 & 13.20 & 13.20 & 13.20 & 13.20 &
      13.20 & 13.20 \\
      \ESD          & 14.29 & 14.04 & 14.55 &
      14.03 & 13.54 & 14.31 & 13.90 & 13.76 &
      13.77 & 15.73 \\
      \Mace         & 13.02 & 13.08 & 13.16 &
      12.82 & 13.05 & 12.74 & 13.09 & 12.95 &
      12.99 & 22.15 \\
      \Receler      & 14.24 & 14.46 & 15.06 &
      13.99 & 13.27 & 14.50 & 13.92 & 13.95 &
      13.81 & 15.88 \\
      \SAFREE       & 27.58 & 23.53 & 26.15 &
      25.99 & 22.05 & 27.70 & 23.14 & 23.71 &
      23.29 & 22.24 \\
      \AdaVD        & 14.37 & 14.37 & 14.35 &
      14.35 & 14.35 & 14.35 & 14.36 & 14.36 &
      14.37 & 14.34 \\
      \UCE          & 13.54 & 13.54 & 13.55 &
      14.20 & 13.98 & 14.01 & 13.32 & 13.74 &
      13.77 & 13.95 \\
      \CURE         & 13.69 & 14.12 & 14.34 &
      13.93 & 14.07 & 13.72 & 13.87 & 13.66 &
      14.07 & 13.78 \\
      \Ours         & 13.52 & 13.51 & 13.65 &
      13.61 & 13.54 & 13.58 & 13.50 & 13.56 &
      13.59 & 14.32 \\
      \bottomrule
    \end{tabular}%
    }
  \end{table}